%% file: MICCAI2026-main_conference_paper_template.tex
\documentclass[runningheads]{llncs}
\usepackage[T1]{fontenc}
\usepackage{graphicx,verbatim}
\usepackage{makecell}
\usepackage{amsmath} 
\usepackage{amssymb}
\usepackage{subcaption}
\usepackage{booktabs}

\begin{document}
\title{MAGE: Color-Invariant and Spatial Knowledge Distillation for Gastric Neoplasm Classification}
%

\author{
Jiho Jun\inst{1,2} \and
Jeongwon Woo\inst{3} \and
Jaemin Song\inst{1,2} \and
Thanh Bong Nguyen\inst{4,2} \and
Dong-heon Yeon\inst{1,2} \and
Donghoon Kang\inst{5} \and
Jae-Myung Park\inst{5} \and
Sung-Jea Ko\inst{1,2} \and
Kwang-Hyun Uhm\inst{6} \thanks{Corresponding author.}
} 


\authorrunning{J. Jun et al.}

\institute{
Korea University, Korea \and
MEDAI, Korea \and
Seoul National University, Korea \and
Vietnam National University, Hanoi, Vietnam \and
The Catholic University of Korea, Seoul ST. Mary’s Hospital Korea \and
Gachon University, Korea\\
    \email{khuhm@gachon.ac.kr}
    }
  
\maketitle              
\begin{abstract}

Accurate differentiation between gastric adenoma and carcinoma during endoscopy is critical for clinical decision-making. Yet, this task is highly challenging due to high inter-class similarity and ambiguous boundaries between the two classes. Existing ROI-based classification methods often suffer from detection/segmentation error propagation and loss of surrounding global context. In contrast, full-image classification lacks the necessary spatial focus. Furthermore, we observe that deep neural networks gravitate towards domain-specific texture biases(e.g. bleeding, lighting artifacts), often causing models to predict based on spurious correlations instead of intrinsic morphological features. To address these limitations, we propose a novel framework, Masked Achromatic Guidance Expert (MAGE). During training, we introduce an auxiliary local expert branch trained on masked achromatic views of the neoplasm. By suppressing background context and color, this branch is forced to learn highly discriminative, purely structural features. We then employ a dual-objective distillation strategy, transferring both classification logits and spatial attention maps to provide implicit spatial supervision to the main branch that receives full WLI as input. This dual-objective distillation forces the model to ground its predictions in morphology rather than relying on shortcuts, while still retaining clinically relevant color cues. At inference time, our deployable model operates on images without annotated masks, ensuring real-time deployability . Extensive experiments on a clinical gastric endoscopy dataset show that our method significantly outperforms existing detection-based methodologies (e.g. YOLO) and classification-based methodologies (e.g. Swin-Transformer), providing not only superior classification performance but also interpretable attention maps for clinical reliability.


\end{abstract}

\section{Introduction}
\input{1_introduction_revision}

\section{Proposed Method}
\input{2_method_new}

\section{Experiments and Results}
\input{3_experiments_and_results_revision}

\section{Conclusion}
\input{4_conclusion}

%
%
%
\bibliographystyle{splncs04}
\bibliography{mybibliography}

\end{document}

%% file: 1_introduction_revision.tex
Distinguishing gastric adenoma from carcinoma in upper gastrointestinal endoscopy is clinically important because it directly affects treatment strategy and prognosis in early gastric neoplasia~\cite{fujishiro2024egc}. 
However, this differential diagnosis remains a challenge even for experts. In a consecutive cohort of superficial elevated gastric neoplasms (n=93), diagnosis using C-WLI achieved 74\% diagnostic accuracy, and implementing M-NBI improved accuracy to 92\%~\cite{maki2013mnbi}. However, this kind of enhanced imaging requires additional resources and could even become substantially harder in clinically ambiguous cases. Furthermore, diagnostic accuracy drops to 55.9\% (C-WLI) / 58.1\% (C-WLI + M-NBI) in deliberately matched difficult cases (50 vs.\ 50; 14 endoscopists)~\cite{tamura2022menbi}. 
These findings highlight the intrinsic difficulty of fine-grained adenoma-carcinoma discrimination, where neoplasms exhibit highly similar appearances and ambiguous boundaries.

Recent deep learning methods for GI endoscopy largely target coarse discrimination (e.g., carcinoma vs.\ background), while fine-grained benign/precancerous vs.\ carcinoma classification remains under-addressed~\cite{liu2026deep}. 
Detection based models show satisfactory performance in localizing neoplasms from background, but have limitations in diagnosis, especially for hard cases~\cite{tudela2024polyp_benchmark}. Conversely, classification models suffer from the black box problem; in our dataset, classification models often predict based on background regions, rather than the ROI itself (see Fig.~\ref{fig:gradcams}). Furthermore, ROI-based pipelines deploying classification models on proposed ROI suffer from localization error propagation, while full-image classification preserves context but can dilute micro-patterns under background and acquisition artifacts~\cite{yin2022endoartefact}.

We hypothesize that the key features distinguishing adenoma from carcinoma are primarily morphological and textural. While color does contain some diagnostic signal, it can also act as a fragile confounder that hinders deep neural networks from focusing on the primary morphological indicators of malignancy. This hypothesis is consistent with magnification-based endoscopic diagnosis literature, where microsurface and microvascular patterns vary with histology and invasion depth~\cite{ok2016menbi_msmv,kurumi2021nbi_review}. In parallel, recent GI imaging studies report that color manipulation/correction do not consistently improve CAD performance and can even increase false positives, suggesting that color information may be unstable or task-dependent~\cite{agossou2025color_correction_capsule}.

Based on this motivation, we propose \textbf{MAGE} (Masked Achromatic Guidance Expert), a framework utilizing a training-time auxiliary expert that learns color-invariant structural cues from a ROI-focused achromatic view (inside-mask grayscale; outside-mask zero). This achromatic expert demonstrated the highest diagnosis performance utilizing ground truth annotated masks, but cannot be deployed at inference time due to the lack of masks. Therefore, we transfer ROI-centric structural knowledge and spatial attention to a deployable model that receives full images as input via soft-target distillation and spatial attention transfer~\cite{hinton2015distill,zagoruyko2017attention}.

In summary, our contributions are threefold: 
(1) a training-only auxiliary expert that promotes color-invariant morphological learning via a masked achromatic input; 
(2) a spatial distillation framework that transfers ROI-centric structural cues to a deployable full-image classifier~\cite{hinton2015distill,zagoruyko2017attention}; and 
(3) consistent improvements over strong detection-and classification-based baselines and domain-pretrained backbones, supported by robustness and explainability analyses.

\begin{figure}[tb]
  \makebox[\textwidth][c]{\includegraphics[width=1\textwidth]{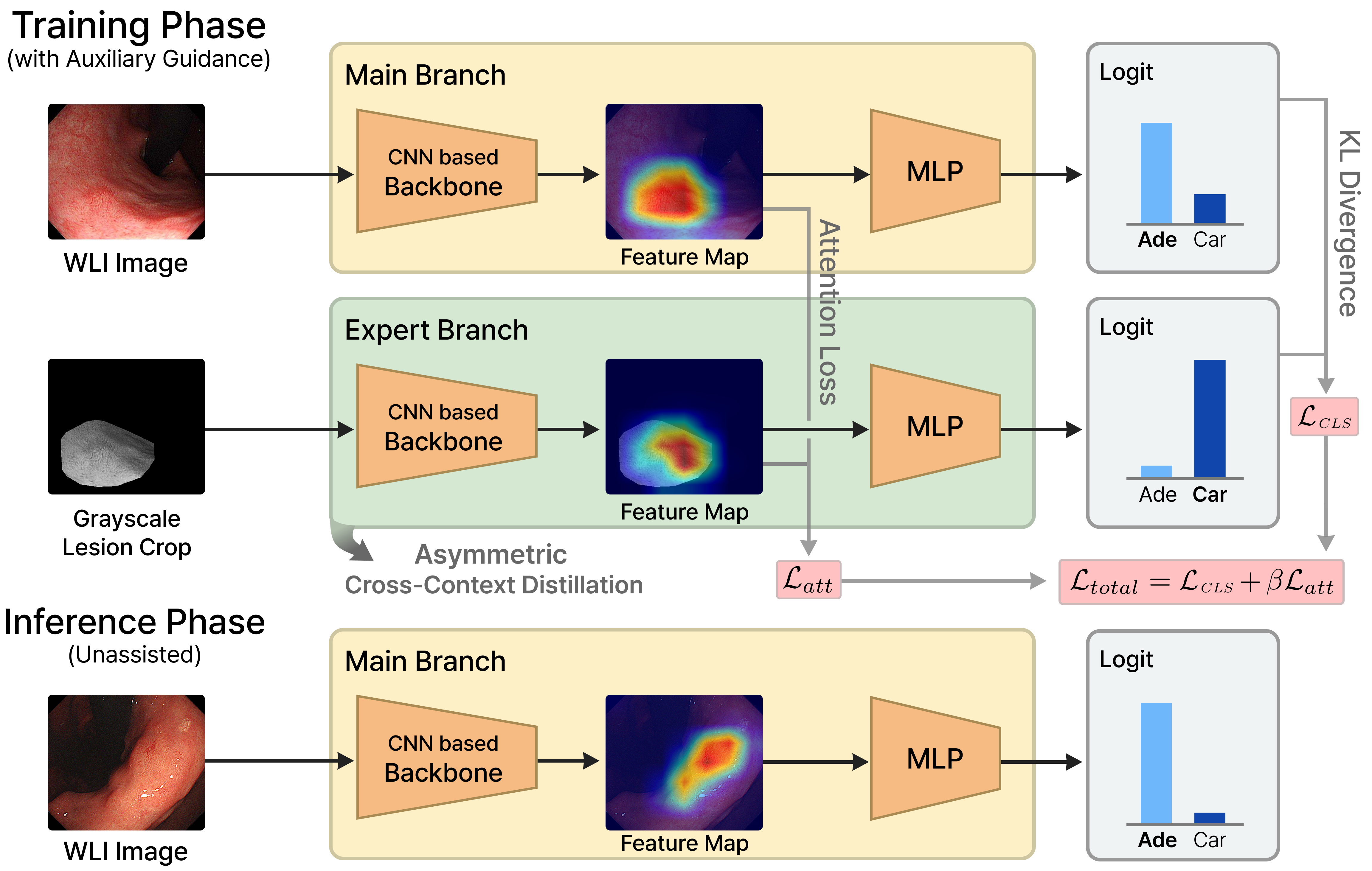}}
  \caption{Overview of the proposed dual-branch training framework.}
  \label{fig:mage_arch}
\end{figure}

%% file: 2_method_new.tex

During the training phase, raw WLI $\mathbf{x}\in\mathbb{R}^{H\times W\times 3}$, annotated masks $\quad
\mathbf{m}\in\{0,1\}^{H\times W}$, and binary labels$\quad
y\in\{0,1\}$ are available. During the inference phase, only the images are available. We define the model's components as the vision encoder $E$, feature map $f = E({\mathbf{x}})$, classification head $C$, and output logit $z = C(f)$.

\subsection{Masked Achromatic Expert}
The Masked Achromatic Expert serves as an auxiliary branch utilizing privileged information to learn color invariant features in the neoplasm area. Because the masked region inherently contains the highest diagnostic signal-to-noise ratio within the image, this constraint forces the expert's feature map to activate within the ROI. Concurrently, training on achromatic views forces the expert to learn robust, color invariant features in a domain where color is often a confounding factor. This training scheme transfers orthogonal training signals to the deployable main model which receives full WLI as input. 
Given a segmentation mask $\mathbf{m}$, we construct a masked-achromatic input
$\tilde{\mathbf{x}}=\mathbf{m}\odot\mathrm{Gray}(\mathbf{x})$,
where $\mathrm{Gray}(\cdot)$ is a luma conversion and the grayscale channel is replicated to 3 channels for the vision encoder. We denote the expert's output logits as $z_E = C_E(E_E(\tilde{\mathbf{x}}))$.

\subsection{Main model}
We transfer knowledge from the expert to the main model via knowledge distillation and spatial attention transfer. 
During training, the expert is kept frozen and produces a feature map and logit from the masked-achromatic input:
$f_E= E_E(\tilde{\mathbf{x}})$ and $z_E = C_E(f_E)$.
In parallel, the main model processes the raw WLI to obtain
$f_M = E_M(\mathbf{x})$ and $z_M = C_M(f_M)$.
Based on these outputs, we optimize the main model with the composite objective
$\mathcal{L}=\mathcal{L}_{\mathrm{CLS}}+\beta\,\mathcal{L}_{\mathrm{att}}$.

\paragraph{Classification Loss}. $\mathcal{L}_{CLS}$ consists of a weighted average between the BCE loss against the ground truth class label($\mathcal{L}_{\mathrm{BCE}}(z_M,y)$) and the KL divergence between the expert's output logits on the image and the model's output logits ($\tau^2\,\mathcal{D}_{\mathrm{KL}}\!\big(p_E^{(\tau)} \,\|\, p_M^{(\tau)}\big))$, where $p_M^{(\tau)}=\sigma\!\left(\frac{z_M}{\tau}\right)$, and $
p_E^{(\tau)}=\sigma\!\left(\frac{z_E}{\tau}\right)$, weighted by $\alpha$ and using temperature $\tau$.

\paragraph{Attention Loss}, $\mathcal{L}_{att}$. In order to transfer spatial attention to the main model, we collapse the channel dimension and normalize the feature map using $A(f) : \mathbb{R}^{B\times C \times H \times W} \xrightarrow{} \mathbb{R}^{B \times H \times W}$. Setting pixels outside the mask to zero restricts expert activations within the ROI. The main model is regularized by squared Frobenius distance  . Using the expert's attention map as a supervision signal implicitly leads the main model to effectively localize neoplasms from the full image. This term is weighted by $\beta$, which controls the strength of the attention regularization.

Together, these terms transfer the expert’s privileged diagnostic knowledge while encouraging neoplasm localization from the full image. Accordingly, the total loss is defined as:

\begin{equation}
\mathcal{L}
=\alpha\,\mathcal{L}_{\mathrm{BCE}}(z_M,y)
+(1-\alpha)\,\tau^2\,\mathcal{D}_{\mathrm{KL}}\!\big(p_E^{(\tau)} \,\|\, p_M^{(\tau)}\big)
+\beta\,\|A(f_M)-A(f_E)\|_F^2 
\end{equation}

\subsection{Rationale for Convolutional Architecture}

While Vision Transformers (ViTs) are widely adopted in endoscopy tasks, we intentionally opted for a Convolutional Neural Network (CNN) backbone. This decision is motivated by dataset constraints and the inherent characteristics of medical imaging. By design, ViTs lack strong inductive biases—such as translation invariance and locality—and therefore require large-scale datasets for effective generalization. Given our dataset, CNN architectures offer crucial regularization via local inductive biases, facilitating stable training without the need for extensive pre-training. 
Furthermore, our proposed distillation framework relies heavily on precise spatial alignment between the intermediate feature maps of the main and expert branches~\cite{zagoruyko2017attention}. The inherent hierarchical, grid-like feature representations of CNNs allow for a more direct and efficient computation of spatial attention loss. Therefore, considering both the visual characteristics of WLI endoscopy and the context-aware knowledge distillation mechanism, our framework is fundamentally tailored for CNN-based architectures.

%% file: 3_experiments_and_results_revision.tex
\subsection{Experimental Setup}

\noindent\textbf{Task.} Binary classification of gastric neoplasia (adenoma vs.\ carcinoma) from WLI.
\textbf{Datasets.} We utilized a private WLI dataset collected from from Seoul St. Mary's Hospital, Korea (IRB: KC20RISI0503), with annotation masks for in-domain training/evaluation, partitioned into patient-disjoint training (840 patients; 2443 images; 1,222 adenoma / 1,221 carcinoma), validation (180 patients; 488 images; 240/248), and test(181 patients; 512 images; 252/260) splits. For external testing, we evaluate our methodology on the PICCOLO~\cite{piccolo2020} dataset(40 patients; 1,727 images; 1,454/ 273), excluding non-neoplastic cases (e.g., hyperplasia).
\textbf{Metrics.} AUC, F1, Precision, and Recall with $\sigma(z_S)$ and a 0.5 threshold unless otherwise stated. Models were trained on an RTX3090 (Adam/AdamW, cosine annealing 1e-4 to 1e-6, batch size 32, 224x224 inputs) utilizing  $\tau=4.0$ and $\beta=300$. All benchmarks used random-flip augmentation only. 

\subsection{Benchmarks}
\newcommand{\mstd}[2]{#1\,{$\pm$\,#2}}

\begin{table}[htb!]
\caption{Main Results (20-run average). \textsuperscript{$\dagger$} Classification-head metrics.}
\label{tab:main_results}
\centering
\small

\begin{tabular}{|>{\raggedright\arraybackslash}p{2.7cm}|c|c|c|c|}
\hline
Method & AUC & F1 & Prec. & Rec. \\
\hline
GastroNet-5M \cite{jong2026gastronet5m} Res50 pretrained
& \mstd{0.744}{0.022} & \mstd{0.704}{0.012} & \mstd{0.683}{0.041} & \mstd{0.734}{0.055} \\
\hline
PolyDSS \cite{saad2024polydss}\textsuperscript{$\dagger$}
& \mstd{0.750}{0.013} & \mstd{0.700}{0.013} & \textbf{\mstd{0.698}{0.015}} & \mstd{0.704}{0.028} \\
\hline
YOLO26-L \cite{yolo26_ultralytics}\textsuperscript{$\dagger$}
& \mstd{0.737}{0.014} & \mstd{0.691}{0.018} & \mstd{0.693}{0.016} & \mstd{0.692}{0.017} \\
\hline
UML \cite{ren_uml_miccai2023}\textsuperscript{$\dagger$}
& \mstd{0.639}{0.038} & \mstd{0.672}{0.012} & \mstd{0.548}{0.028} & \mstd{0.879}{0.080} \\
\hline
SAM (ViT-B) \cite{kirillov2023sam}+\linebreak
Swin-Tiny\cite{liu2021swin}
& \mstd{0.719}{0.020} & \mstd{0.700}{0.014} & \mstd{0.586}{0.035} & \textbf{\mstd{0.879}{0.061}} \\
\hline
U-Net (Res50)+ Swin-Tiny \cite{liu2021swin}
& \mstd{0.731}{0.012} & \mstd{0.718}{0.011} & \mstd{0.615}{0.036} & \mstd{0.870}{0.060} \\
\hline
\textbf{Ours (MAGE, $\alpha=0.25$)}
& \textbf{\mstd{0.786}{0.010}} & \textbf{\mstd{0.733}{0.010}} & \mstd{0.673}{0.017} & \mstd{0.806}{0.035} \\
\hline
\end{tabular}
\end{table}

Table~\ref{tab:main_results} compares our approach with representative baselines and recent Gastronet 5M pretrained backbones.
To ensure fair comparison across heterogeneous architectures, we report image-level classification metrics: joint models use their classification head outputs, and two-stage pipelines(e.g. SAM + Swin-Transformer) classify ROIs localized by segmentation. We report the average and standard deviation across 20 runs. In particular, MAGE improves F1 and AUC over the strongest classification baselines, indicating more reliable overall decision quality under the same WLI-only setting.

\subsection{Explainability and Robustness}

\subsubsection{Explainability: Grad-CAM / Attention Concentration}
\begin{figure}[htb!]
  \makebox[\textwidth][c]{\includegraphics[width=0.6\textwidth]{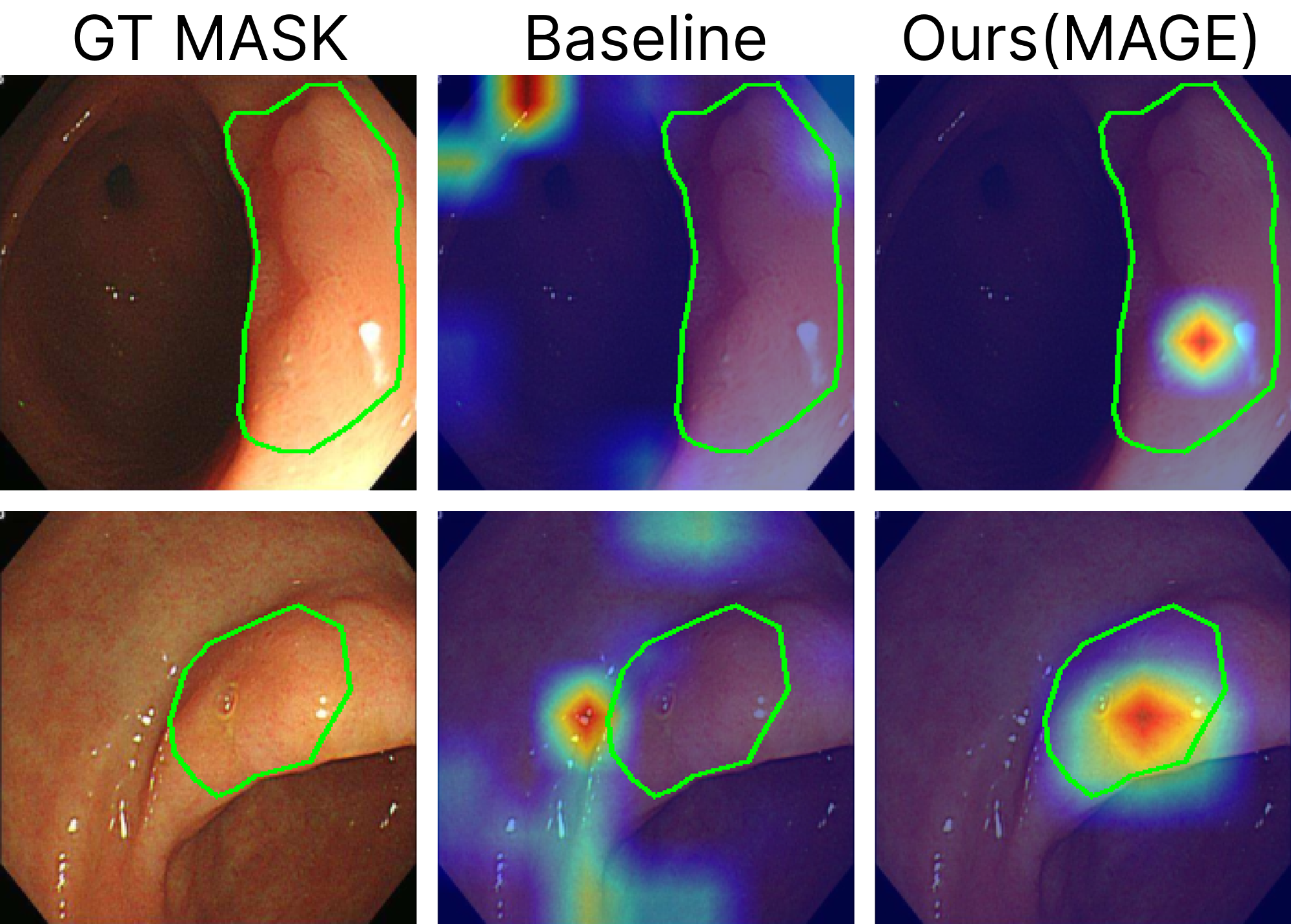}}
  \caption{Examples of Grad-CAM, GT, Baseline(EfficientNet), MAGE.}
  \label{fig:gradcams}
\end{figure}

We analyze model explainability using Grad-CAM~\cite{selvaraju2017gradcam} and quantify how well the resulting attribution maps align with neoplasm masks (Table~\ref{tab:gradcam}).
Specifically, we report: (i) \emph{Attention-in-Mask (AiM)}, the fraction of total attribution mass that falls inside the ground-truth mask; (ii) the \emph{Pointing Game Accuracy}(PGA) ~\cite{zhang2016excitation_backprop}, which checks whether the maximum-saliency point lies within the neoplasm region; and (iii) a \emph{saliency Dice} score computed between binarized saliency maps and the ground-truth masks, where the binarization threshold is selected on the validation set to maximize the score. Overall, MAGE yields more localized and mask-consistent explanations, indicating that the model relies more on neoplasm morphology/structure rather than spurious global color patterns. Notably, its explanation quality is comparable to that of a strong baseline trained directly on segmentation objectives. Fig.~\ref{fig:gradcams} presents qualitative Grad-CAM examples.  

\begin{table}[htb!]
\caption{Grad-CAM + Attention Evaluation. \textsuperscript{$\dagger$} YOLO reports standard segmentation Dice (not saliency mDice) and is shown for reference.}
\label{tab:gradcam}
\centering
\begin{tabular}{|l|l|l|l|}
\hline
Method & AiM & PGA & Saliency mDice \\
\hline
Baseline (EffNet) & 0.3176 & 0.4355 & 0.3225 @ $t=0.15$ \\
GastroNet pretrained ResNet50 & 0.3727 & 0.5117 & 0.3600 @ $t=0.10$ \\
Swin-Transformer Tiny & 0.2023 & 0.3398 & 0.2492 @ $t=0.10$ \\
\hline
Ours ($\alpha=0.5$) & 0.4876 & 0.6758 & 0.4751 @ $t=0.20$ \\
Ours ($\alpha=0.25$) & 0.5303 & 0.7324 & 0.5308 @ $t=0.20$ \\
\textbf{Ours ($\alpha=0$)} & \textbf{0.5484} & \textbf{0.7734} & \textbf{0.5489 @ $t=0.20$} \\
\hline
YOLO26-Seg Large\textsuperscript{$\dagger$} (\textit{for reference}) & 0.6313 & 0.7695 & 0.6361 (Dice) \\
\hline
\end{tabular}
\end{table}

\subsubsection{Color Robustness: Hue Shift / Color Perturbation}

\begin{figure}[htb!]
  \centering
  \begin{subfigure}[c]{0.49\linewidth}
    \centering
    \includegraphics[width=\linewidth,height=0.27\textheight,keepaspectratio]{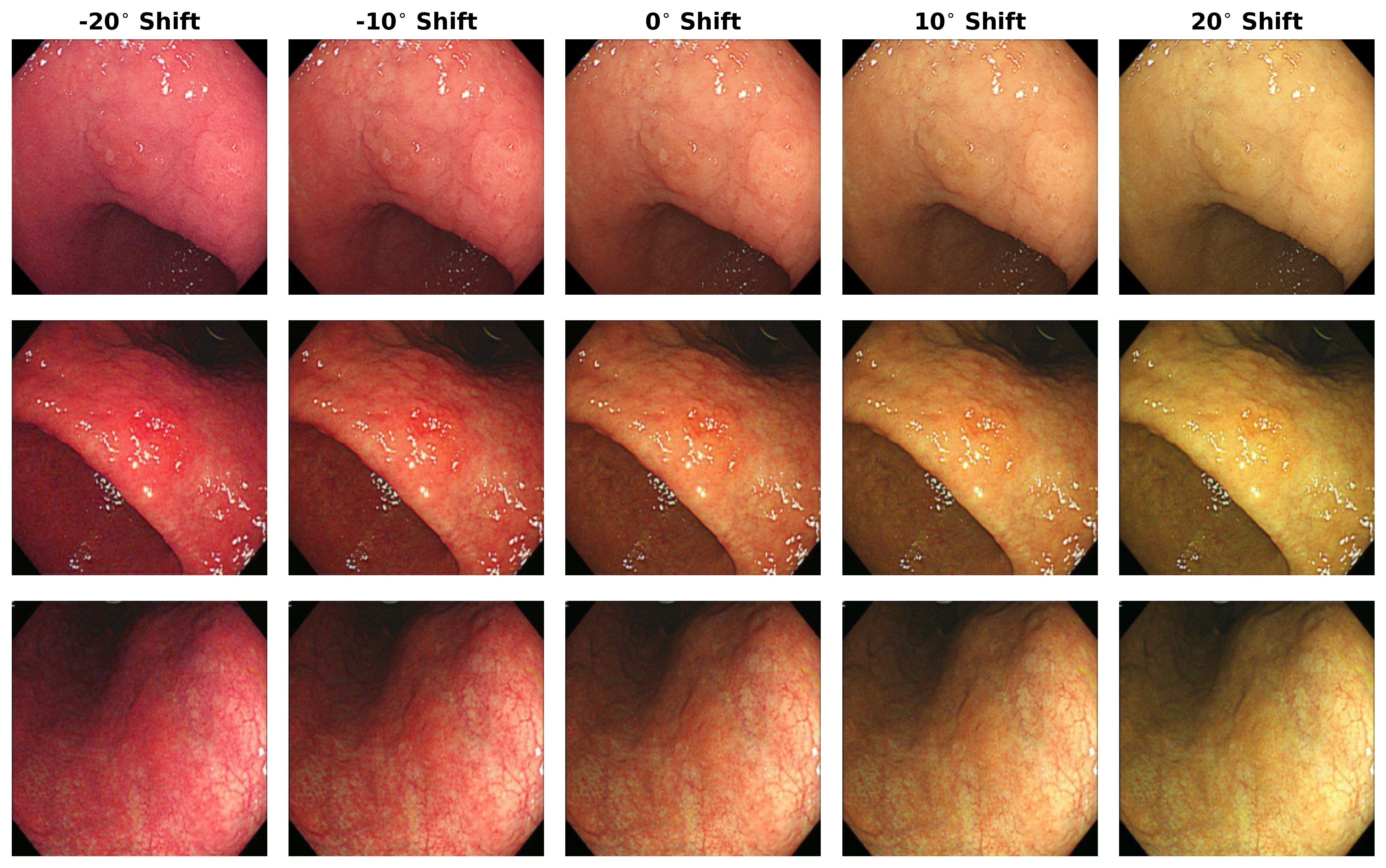}
    \caption{Examples of hue shifts ($-20^\circ$ to $+20^\circ$).}
    \label{fig:hue_examples}
  \end{subfigure}\hfill
  \begin{subfigure}[c]{0.49\linewidth}
    \centering
   \includegraphics[width=\linewidth,height=0.27\textheight,keepaspectratio]{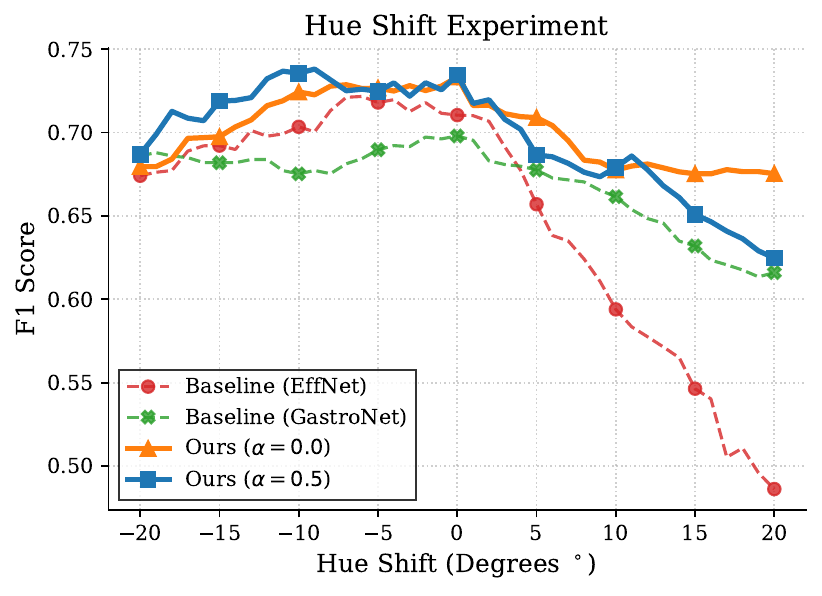}
    \caption{Robustness comparison (MAGE vs.\ baseline).}
    \label{fig:hue_curve}
  \end{subfigure}

  \caption{Color robustness under synthetic hue perturbations.}
  \label{fig:hue_shift}
\end{figure}

Gastric endoscopy images often exhibit non-diagnostic appearance variations that can shift the observed color tone and illumination.
To test the model's robustness against color variations, we evaluate performance on images subject to synthetic hue rotations.
Specifically, we apply a range of hue shifts to the validation set and compare our model against a baseline (EffNet-v2-s) and benchmark model(ResNet50 pretrained on GastroNet-5M, finetuned on our dataset). As shown in Fig.~\ref{fig:hue_shift}, our method shows superior overall performance compared to both the EfficientNet trained baseline and GastroNet-5M pretrained ResNet50 baseline. Especially, while the naively trained EfficientNet degrades significantly over positive hue shifts, our MAGE model ($\alpha=0$) shows robustness on par with the GastroNet pretrained baseline. Although ($\alpha=0$) maximizes color robustness and spatial alignment, we utilize ($\alpha=0.25$) in benchmarks to optimally balance label-guided accuracy.

\subsection{External Dataset: PICCOLO (Cross-domain Generalization)}

\begin{table}[htb!]
\caption{PICCOLO results.}
\label{tab:piccolo}
\centering
\small
\begin{tabular}{|l|l|l|l|l|}
\hline
Method & AUC & mF1 & mPrecision & mRecall \\
\hline
\makecell[l]{GastroNet-5M pretrained~\cite{jong2026gastronet5m} ResNet50}
& \makecell[c]{0.9110}
& \makecell[c]{0.8053}
& \makecell[c]{0.8269}
& \makecell[c]{0.6717} \\
\hline
\makecell[l]{Swin-Tiny~\cite{liu2021swin} (ImageNet pretrained)}
& \makecell[c]{0.9293}
& \makecell[c]{\textbf{0.8534}}
& \makecell[c]{0.9004}
& \makecell[c]{\textbf{0.8440}} \\
\hline
\makecell[l]{U-Net (Res50) + Swin-Tiny~\cite{liu2021swin}}
& \makecell[c]{0.9096}
& \makecell[c]{0.7858}
& \makecell[c]{0.7975}
& \makecell[c]{0.8099} \\
\hline
\makecell[l]{YOLO 26 Large~\cite{yolo26_ultralytics}}
& \makecell[c]{0.7816}
& \makecell[c]{0.7521}
& \makecell[c]{0.8566}
& \makecell[c]{0.6959} \\
\hline
\makecell[l]{Ours (MAGE)}
& \makecell[c]{\textbf{0.9434}}
& \makecell[c]{0.8173}
& \makecell[c]{\textbf{0.9298}}
& \makecell[c]{0.7863} \\
\hline
\end{tabular}
\end{table}

We further evaluate cross-domain generalization on PICCOLO~\cite{piccolo2020}, an external colon endoscopy dataset. 
We test whether our training methodology—focused on color-invariant, neoplasm-centric morphological features—is transferable beyond the gastric domain. Due to the limited number of unique neoplasms in PICCOLO, we report metrics averaged over a 10-repetition 3-fold cross validation.
We report AUC and macro-averaged F1/Precision/Recall to account for class imbalance.
As shown in Table~\ref{tab:piccolo}, our method achieves the highest AUC, supporting the overall robustness under the change of the domain.

\subsection{Ablation Study}

The highest performance was achieved by pairing a cropped grayscale Expert, which focuses on intrinsic neoplasm structure, with an RGB Main branch that leverages morphological and color cues. The Main branch exhibits a limited ability to distinguish neoplasms in grayscale images. However, for the expert receiving masked inputs, training on grayscale inputs exhibited a higher classification performance compared to RGB inputs. These findings suggest that an \emph{asymmetric design} intentionally constrains spatial and color information to force the model to learn structural, color-invariant features. $\alpha=0.25$ was used in training the main inputs.

\begin{table}[htb!]
\caption{Ablation Study on Knowledge Distillation Configurations.}\label{tab:ablation_study}
\centering
\footnotesize

\begin{tabular}{|c|c|c|c|c|c|c|c|}
\hline
\multicolumn{4}{|c|}{\textbf{Expert Configuration}} & \multicolumn{2}{c|}{\textbf{Main Input : Grayscale}} & \multicolumn{2}{c|}{\textbf{Main Input : RGB}} \\
\hline
\textbf{RGB} & \textbf{Gray} & \textbf{Full} & \textbf{Crop} & \textbf{AUC} & \textbf{F1} & \textbf{AUC} & \textbf{F1} \\
\hline
\checkmark & & \checkmark & & 0.7483 & 0.7081 & 0.7738 & 0.7192 \\
\checkmark & & & \checkmark & 0.7082 & 0.6872 & 0.7748 & 0.7177 \\
& \checkmark & \checkmark & & 0.7524 & 0.7147 & 0.7767 & 0.7235 \\
& \checkmark & & \checkmark & 0.7115 & 0.6940 & \textbf{0.7860} & \textbf{0.7330} \\
\hline
\end{tabular}
\end{table}

%% file: 4_conclusion.tex
\noindent\textbf{Contribution.}
We presented MAGE, a dual-branch training framework that promotes color-invariant morphological representations for gastric neoplasm diagnosis.
Our key contribution is a training-only masked grayscale expert that suppresses color shortcuts, and distills its structural knowledge into a deployable full-image model.
As a result, MAGE improves classification performance and robustness under color/domain variability.

\noindent\textbf{Limitations and Future Work.}
Our study focuses on classification and robustness without explicitly training a dedicated detection/segmentation module, which may limit direct localization outputs required in some clinical workflows. A natural next step is to leverage the learned classifier for lightweight localization (e.g., CAM-based or weakly supervised cues) without incurring the cost of a full segmentation system.
In addition, broader external validation, particularly multi-center and device-diverse cohorts, remains essential to establish generalization and clinical utility. Code and weights will be publicly released.